\documentclass{article}
\usepackage{kr}

\usepackage{times}
\usepackage{url}
\usepackage[hidelinks]{hyperref}
\usepackage[utf8]{inputenc}
\usepackage[small]{caption}
\usepackage{graphicx}
\usepackage{amsmath}
\usepackage{amsthm}
\usepackage{booktabs}
\usepackage{algorithm}
\usepackage{algorithmicx}
\usepackage[english]{babel}
\usepackage[letterpaper,top=2cm,bottom=2cm,left=3cm,right=3cm,marginparwidth=1.75cm]{geometry}
\usepackage{mathtools}
\usepackage[all,cmtip]{xy}
\usepackage{etoolbox}

\newtheorem{example}{Example}
\newtheorem{theorem}{Theorem}

\title{Approximation Fixpoint Theory with Refined Approximation Spaces}

\author{%
	Linde Vanbesien$^{1, 2}$\and
	Bart Bogaerts$^2$\and
	Marc Denecker$^1$\\
	\affiliations
	$^1$Department Of Computer Science, KU Leuven, Belgium\\
	$^2$Department Of Computer Science, Vrije Universiteit Brussel, Belgium\\
	\emails
	\{linde.vanbesien, marc.denecker\}@kuleuven.be,
	bart.bogaerts@vub.be,
}

\newbool{beamer}
\newbool{article}
\booltrue{article}
\usepackage[usenames,dvipsnames]{xcolor}
\usepackage{times}
\usepackage[utf8]{inputenc}

\usepackage{tikz}
\usepackage{url}
\usepackage{amsthm}
\usepackage{amssymb}
\usepackage{amsmath}
\usepackage{paralist}
\usepackage{xspace}
\usepackage[capitalize]{cleveref}
\usepackage{centernot}
\usepackage{mathtools}
\usepackage{stmaryrd}
\usepackage{array,  makecell}
\usepackage{xparse}
\usepackage{stackengine}
\usepackage{fancybox}
\usepackage{subcaption}
\usepackage{multicol}
\usepackage{bm}

\usepackage{algorithm}
\usepackage[noend]{algpseudocode}  

\usetikzlibrary{positioning}
\tikzset{main node/.style={circle,fill=white!20,draw,minimum size=0.4cm,inner sep=0pt},}
\tikzset{old node/.style={circle,fill=white!20,draw=lightgray,minimum size=0.4cm,inner sep=0pt,text=lightgray},}
\tikzset{dis node/.style={text=lightgray},}
\tikzset{new edge/.style={red,dotted}}
\tikzset{weak edge/.style={lightgray}}
\tikzset{weak op/.style={lightgray, thick}}
\tikzset{new node/.style={circle,fill=red!30,draw=red,minimum size=0.4cm,inner sep=0pt,dotted}}
\tikzset{inc node/.style={draw=blue,fill=blue!30,minimum size=0.4cm,inner sep=0pt, rounded corners}}
\tikzset{bound node/.style={draw=green,fill=green!30,minimum size=0.4cm,inner sep=0pt, rounded corners}}
\tikzset{upbound node/.style={draw=red,fill=red!30,minimum size=0.4cm,inner sep=0pt, rounded corners}}
\tikzset{shaded node/.style={fill=black!20}}
\usetikzlibrary{calc}

\newcommand\citet[1]{\citeauthor{#1}~(\citeyear{#1})\xspace}

\newcommand{\itz}{\begin{itemize}\item}
\newcommand{\eitz}{\end{itemize}}

\newcommand{\m}[1]{\ensuremath{#1}\xspace}
\newcommand\ie{i.e.,\xspace}

\newcommand{\ignore}[1]{}

\newboolean{nocomments}
\setboolean{nocomments}{false}

\newcommand{\showcomment}[1]{\ifthenelse{\boolean{nocomments}}{\relax}{#1}}
\newcommand{\showbeamer}[1]{\ifbool{beamer}{\relax}{#1}}
\newcommand{\showarticle}[1]{\ifbool{article}{\relax}{#1}}

\newboolean{journal}
\setboolean{journal}{false}

\newcommand{\conf}[1]{\ifthenelse{\boolean{journal}}{\relax}{#1}}
\newcommand{\journ}[1]{\ifthenelse{\boolean{journal}}{#1}{\relax}}

\newcommand{\restr}[2]{{
  \left.\kern-\nulldelimiterspace 
  #1 
  \vphantom{\big|} 
  \right|_{#2} 
}}

\newcommand{\AFT}{AFT\xspace}

\algnewcommand{\IfThenElse}[3]{
\State \algorithmicif\ #1\ \algorithmicthen\ #2\ \algorithmicelse\ #3}

\newcommand\Rules[1][ ]{%
	\if\relax\detokenize{#1}\relax
	\m{\mathcal{R}}%
	\else
	\m{\mathcal{R}_{\!#1}}%
	\fi
}
\newcommand\ES[1][ ]{%
	\if\relax\detokenize{#1}\relax
	\m{\mathcal{E}}%
	\else
	\m{\mathcal{E}_{\!#1}}%
	\fi
}
\newcommand\CS[1][ ]{%
	\if\relax\detokenize{#1}\relax
	\m{\mathcal{C}}%
	\else
	\m{\mathcal{C}_{#1}}%
	\fi
}

\newcommand\FS[1][\CS]{%
	\if\relax\detokenize{#1}\relax
	\m{\mathcal{F}}%
	\else
	\m{\mathcal{F}^{\!#1}}%
	\fi
}

\newcommand\AF[1][ ]{%
	\if\relax\detokenize{#1}\relax
	\m{\mathcal{A}}%
	\else
	\m{\mathcal{A}_{#1}}%
	\fi
}

\newcommand\AbS[1][ ]{%
	\if\relax\detokenize{#1}\relax
	\m{\mathcal{B}}%
	\else
	\m{\mathcal{B}_{#1}}%
	\fi
}

\newcommand{\lowlat}[1][ ]{\m{\mathcal{L}_{#1}}}
\newcommand{\uplat}[1][ ]{\m{\mathcal{U}_{#1}}}

\newcommand\cpo[1][ ]{%
	\if\relax\detokenize{#1}\relax
	\m{\set{C}}%
	\else%
	\m{\set{C}_{\!#1}}%
	\fi%
}

\newcommand\flower[1][]{%
	\if\relax\detokenize{#1}\relax
	\m{\set{F}}%
	\else%
	\m{\set{F}_{\!#1}}%
	\fi%
}

\newcommand\convex[1][]{%
	\if\relax\detokenize{#1}\relax
	\m{CC}%
	\else%
	\m{CC_{\!#1}}%
	\fi%
}

\newcommand\imcons{\m{O}}

\newcommand{\ult}[1][ ]{\m{U_{#1}}}
\newcommand{\apimcons}[1][ ]{\m{\mathcal{\imcons}}}

\newcommand{\leqp}{\m{\leq_p}}
\newcommand{\leqt}{\m{\leq_t}}

\newcommand{\leqa}{\m{\leq_a}}
\newcommand{\leqA}{\m{\preceq}}

\newcommand{\leqf}{\m{\leqA_f}}

\makeatletter
\newcommand{\oleqx}[1][$\leqA$]{\mathbin{\mathpalette\make@circled{\raisebox{0.1 em}{\scalebox{0.5}{#1}}}}}
\newcommand{\make@circled}[2]{%
	\ooalign{$\m@th#1\bigcirc{#1}$\cr\hidewidth$\m@th#1#2$\hidewidth\cr}%
}

\newcommand{\aop}[1][]{
	\if\relax\detokenize{#1}\relax
	\m{A}%
	\else
	\m{A_{#1}}
	\fi
}

\newcommand{\appr}[1][\AbS]{\m{\mathit{Appr}({#1})}}

\newcommand{\aoplx}[1]{\m{\aop[#1_2]^1}}
\newcommand{\aopux}[1]{\m{\aop[#1_1]^2}}
\newcommand{\aopllx}[1]{\m{\low[{\aop[{\up[{#1}]}]}]}}
\newcommand{\aopuux}[1]{\m{\up[{\aop[{\low[{#1}]}]}]}}

\newcommand{\sto}[1][]{%
        \if\relax\detokenize{#1}\relax
        \m{\mathit{St}}%
        \else
        \m{\mathit{St}_{\!#1}}
        \fi
}

\newcommand{\lc}[1]{\m{{#1}^\downarrow}}
\newcommand{\uc}[1]{\m{{#1}^\uparrow}}
\newcommand{\stl}[1][]{\m{{\sto[{#1}]}^1}}
\newcommand{\stu}[1][]{\m{{\sto[{#1}]}^2}}

\newcommand{\Min}[1]{\m{\mathit{Min}(#1)}}
\newcommand{\Max}[1]{\m{\mathit{Max}(#1)}}

\newcommand{\flowmin}[1][\CS]{\m{\lowlat[f]}}
\newcommand{\flowmax}[1][\CS]{\m{\uplat[f]}}
\newcommand{\ccmin}[1][\CS]{\m{\lowlat[cc]}}
\newcommand{\ccmax}[1][\CS]{\m{\uplat[cc]}}

\newcommand{\gal}[1][]{\m{\iota_{#1}}}

\newcommand{\set}[1]{\m{\boldsymbol{#1}}}

\newcommand{\low}[1][.]{\m{#1^{\lowlat}}}
\newcommand{\up}[1][.]{\m{#1^{\uplat}}}

\newcommand{\alb}{ALB\xspace}
\newcommand{\aub}{AUB\xspace}
\newcommand{\albs}{ALBs\xspace}
\newcommand{\aubs}{AUBs\xspace}
\newcommand{\AbstractILP}{Abstract Interlattice Lub Property\xspace}
\newcommand{\ChainILP}{Chain Interlattice Lub Property\xspace}
\newcommand{\WeakILP}{Weak Interlattice Lub Property\xspace}
\newcommand{\IGP}{Interlattice Glb Property\xspace}

\newcommand{\KK}{\m{\mathit{K\!\!K}\!}}
\newcommand{\WF}{\m{\mathit{W\!\!F}\!}}
\newcommand{\ST}{\m{\mathit{S\!\!T}\!}}
\newcommand{\SUP}{\m{\mathit{S\!U\!\!P}\!}}

\DeclareMathOperator{\glb}{glb}
\DeclareMathOperator{\lub}{lub}

\DeclareMathOperator{\lfp}{lfp}
\DeclareMathOperator{\gfp}{gfp}

\showbeamer{\newtheorem{definition}{Definition}
\showarticle{\newtheorem{theorem}{Theorem}
\newtheorem{example}{Example}}
}
\newtheorem{prop}{Proposition}

\newtheorem{require}{Requirement}

\begin{document}
	\maketitle
	
	\begin{abstract}
		Approximation Fixpoint Theory (\AFT) is a powerful theory covering various semantics of non-monotonic reasoning formalisms in knowledge representation such as Logic Programming and Answer Set Programming.  Many semantics of such non-monotonic formalisms can be characterized as suitable fixpoints of a non-monotonic operator on a suitable lattice. Instead of working on the original lattice, \AFT operates on \emph{intervals} in such lattice to approximate or construct the fixpoints of interest. While \AFT has been applied successfully across a broad range of non-monotonic reasoning formalisms, it is confronted by its limitations in other, relatively simple, examples. In this paper, we overcome those limitations by extending consistent AFT to deal with approximations that are more refined than intervals. Therefore, we introduce a more general notion of approximation spaces, showcase the improved expressiveness and investigate relations between different approximation spaces.
	\end{abstract}
	
	\section{Introduction}
	Approximation Fixpoint Theory (\AFT), introduced by \citet{DeneckerMT00}, is a comprehensive constructive framework covering various semantics for non-monotonic reasoning formalisms in knowledge representation. In such reasoning formalisms, it is common practice and often very natural to associate a \emph{semantic operator} on a lattice (\emph{the exact lattice}) with a theory (or program). In case this operator is monotonic, the semantics will typically coincide with the least fixpoint of the operator. However, in general, the semantic operator is not monotonic. This obfuscates the link between the intended semantics of the theory and the semantic operator. In fact, for many non-monotonic reasoning formalisms, different (disagreeing) semantics exist. This is where \AFT comes in. Instead of the semantic operator, it uses an \emph{approximating operator} and reduces different proposed semantics to different types of fixpoints of this approximating operator.
	As the name suggests, the approximating operator operates on an \emph{approximation space}, \ie it maps approximations of exact elements to other approximations. Essentially, an approximating operator uses approximations to approximate the behavior of the semantic operator. Following its original introduction, \AFT uses \emph{pairs} of elements in the exact space as approximations. Such pairs can be thought of as intervals, therefore we will refer to this version of \AFT as interval-\AFT.
	The semantics captured by \AFT include the supported, Kripke-Kleene, stable and the well-founded semantics.
	 It has been shown that \AFT is able to grasp multiple non-monotonic reasoning formalisms, including (higher order) Logic Programming \cite{iclp/DeneckerPB01,tplp/CharalambidisRS18}, 
auto-epistemic and default logic \cite{DeneckerMT03}, and argumentation frameworks and abstract dialectal frameworks \cite{journals/ai/Strass13,aaai/Bogaerts19}. 

\vspace{-0.2cm}
\paragraph{Extending Consistent AFT} 
In this paper, we develop a generalization of consistent AFT in which the elements of the approximation space need not necessarily be pairs/intervals, but more general approximations are allowed. Our new theory is a true generalization of consistent AFT: when instantiated with pairs as approximations, it reduces to the standard definitions of \citet{DeneckerMT04}. Our generalization is motivated by examples from auto-epistemic reasoning and abstract argumentation (more specifically weighted abstract dialectic frameworks \cite{aaai/BrewkaSWW18}, which we discuss next.

\vspace{-0.2cm}
\paragraph{Motivating Example from AEL} 
It has been argued that logic programming under well-founded model semantics can be understood as a logic of inductive definitions \cite{KR/DeneckerV14}. In this, \AFT's well-founded fixpoint construction constructs the relations inductively defined by the program. \citet{KR/DeneckerV14} claim that for such definitional programs, the well-founded model will always identify the defined set. In the application of \AFT for auto-epistemic logic (AEL), the well-founded fixpoint construction does not construct sets, but a belief state. 
	\begin{example}[First published by \citet{tocl/BogaertsVD16}]\label{ex:Hanne}
		Consider the following 
		auto-epistemic theory:
		\[\mathcal{T} = \left\{\begin{array}{c} \mathit{q} \Leftrightarrow \neg K \mathit{p}.\\\mathit{r} \Leftrightarrow \neg K \mathit{q}.\end{array}\right\}.\]
	\end{example}
	This theory is seen as the theory of a rational, epistemic agent with introspection, whose knowledge base does not only contain objective facts of the world, but also propositions about what it knows or does not know. E.g., suppose the above theory would contain the objective fact `$p.$'  then the agent would know $p$, $Kp$ would be true, and according to the agents theory, $q$ would be false, which in turn would lead to $r$ being true according to the theory. But as it happens, the theory does not contain objective information about $p$, hence the agent does not know $p$, meaning that $Kp$ is false, and hence $q$ is true, and further down the road, $r$ is false. This is the belief state of the agent: nothing is known about $p$, but $q$ and $\neg r$ are known to be true. This at least is the intended belief state of this theory.
	
	\citet{tocl/BogaertsVD16} explain how the well-founded semantics fails for this example. In fact, it is not able to derive any information. 
	The well-founded model is the least precise approximation, \ie the approximation that approximates all considered interpretations. Therefore, the authors stated that \AFT is not strong enough for some auto-epistemic theories and they proposed an alternative method to determine the semantics of those theories using the concept of well-founded sets. 
	This is a powerful ad-hoc algebraic method that is able to find the expected model of theories of this kind. 
	Intuitively, in that paper, arbitrary sets of elements are used as approximations and a powerful algebraic method is defined that can identify the intuitive model in (a class of theories generalizing) the above example. 
	In this paper, on the other hand, we show that the \emph{standard} methods of AFT suffice, provided we generalize them to a richer approximation space. 
	
%
	\vspace{-0.4cm}
	\paragraph{Motivating Example from wADFs} 
	Likewise, \citet{aaai/Bogaerts19} discusses how \AFT can be used to successfully capture a broad range of weighted abstract dialectical frameworks (wADFs), which were introduced by \citet{aaai/BrewkaSWW18}. In essence, wADFs are systems that represent knowledge on how different arguments influence each other, \ie arguments can attack and support one another, possibly by joint attacks or support. An interpretation of a wADF then assigns an \emph{acceptance value} to every argument which represents the degree to which an argument is accepted. Naturally, this requires an acceptance order $\leqa$ on the set of possible acceptance values. Due to the fact that \AFT uses intervals as approximations, \citet{aaai/Bogaerts19} had to pose restrictions on the acceptance order. In particular, in order to have a least precise approximation, there needs to be a `least precise' upper bound, \ie a greatest element with respect to the acceptance order. On the other hand, \citet{aaai/BrewkaSWW18} allows multiple $\leqa$-maximal acceptance values. In other words, interval-\AFT requires the exact space to be a complete lattice, whereas not every problem naturally satisfies this constraint.
	
	\begin{example}[Corresponding to Example 3.2 as introduced by \citet{aaai/BrewkaSWW18}]\label{ex:Paper}
		Consider the following wADF:
		\begin{figure}[H]
			\centering
			\vspace{-0.4cm}
			\begin{subfigure}{\linewidth}
				\begin{tikzpicture}[node distance=0.2cm]
					\node (status) at (2, 1) {paper's status};
					\node (sign) at (0, 0) {paper's significance};
					\node (meth) at (4, 0) {scientific methodology};
					
					\path[draw,->]
					(sign) edge node {} (status)
					(meth) edge node {} (status);	
				\end{tikzpicture}
				\caption{A simple wADF illustrating how the status of a paper depends on its significance and the used methodology.}
			\end{subfigure}
			\begin{subfigure}{\linewidth}
				\begin{tikzpicture}[node distance=0.2cm]					
					\node (acc) at (0, 2) {accept};
					\node (bord) at (3, 2) {borderline};
					\node (rej) at (6, 2) {reject};
					\node (tacc) at (1.5, 1) {tendency accept};
					\node (trej) at (4.5, 1) {tendency reject};
					\node (u) at (3, 0) {indifferent};
					
					\path[draw,-]
					(acc) edge node {} (tacc)
					(bord) edge node {} (tacc)
					(bord) edge node {} (trej)
					(rej) edge node {} (trej)
					(tacc) edge node {} (u)
					(trej) edge node {} (u);	
				\end{tikzpicture}
				\caption{The acceptance values under $\leqa$.}
			\end{subfigure}
		\end{figure}
	\end{example}
	The wADF states that the status of a paper is jointly supported by its significance and the used scientific methodology. Therefore, the value for the status of the paper will be the greatest value that is compatible with the values of the significance and the methodology, \ie the greatest lower bound of those values. For example, let us assume that the significance of the paper has the value `accept'. On the other hand, peer reviewing determined that the scientific methodology should be assigned the value `borderline'. In this case, the intended model assigns the value `tendency accept' to the paper's status. The order $\leqa$ can be interpreted as the confidence level. For example, while both `borderline' and `indifferent' are essentially neutral opinions, they are very different in nature. When an objective reviewer starts reviewing a paper, he will be `indifferent' to the paper, he does not have enough information yet to give a strong opinion. On the other hand, by the time the reviewer thoroughly read through the paper, he might be of the opinion that the paper is a `borderline' paper. While a reviewer with an `indifferent' opinion will most likely not disagree with someone who wants to nominate the paper for an award, the reviewer with the opinion `borderline' definitely will.
	Clearly the set of acceptance values under $\leqa$ lacks a greatest element, therefore, \AFT is not able to correctly represent the least precise approximation. 
	
	\vspace{-0.4cm}
	\paragraph{Our Contributions}
	The two examples we discussed above demonstrate some of the limitations of standard \AFT with intervals. In order to resolve these limitations, this paper introduces a more flexible notion of the approximation space enabling the use of more refined approximations. On the one hand, more precise approximations naturally hold more information and therefore, they may result in stronger expressiveness needed to solve, e.g., \cref{ex:Hanne}. On the other hand, allowing more flexible approximation spaces extends the usability of \AFT to formalisms that are inherently more flexible, such as wADFs. The exact space in \cref{ex:Paper} seems to require a set of multiple `maximal elements' as upper bound. 
	Therefore, in \cref{sect:Flowers}, we take a closer look at \emph{flowers} which have more refined upper bounds. We identify a set of interesting and desirable properties of this space which serve as inspiration for the more relaxed definition of an approximation space given in \cref{sect:Theory}. We show that our extension of \AFT is conservative in upholding the core intuitions and theorems of consistent interval-\AFT as published by \citet{DeneckerMT04}, while concurrently expanding the theory's scope by introducing new approximation spaces such as the space of flowers. Using flowers as approximations, \AFT is capable of deriving the intended well-founded model for \cref{ex:Hanne}. Additionally, flower-based \AFT facilitates a straightforward extension of the work by \cite{aaai/Bogaerts19} towards any wADFs such as \cref{ex:Paper}. Consequently, the proposed generalization of \AFT successfully increases the expressiveness while simultaneously introducing more flexibility. Furthermore, the generalization brings about a deeper understanding of \AFT by zooming in on the necessary conditions. 
	On the other hand, the generalization also introduces a new challenge. Whereas, given a problem in an exact space, the approximation space used to be a fixed entity derivable from the exact space, now an exact space may lead to multiple suitable approximation spaces. This brings up the question of how to select an appropriate approximation space. In \cref{sect:Hierarchy}, we investigate the relation between approximation spaces within the problem environment. We define a \emph{precision-order} on the approximation spaces such that more precise approximation spaces constitute more accurate semantics provided we use appropriate approximating operators. On the other side of the coin, more precise approximation spaces in general introduce higher computational complexity. Thus, a good approximation space strikes a good balance between computational complexity and expressiveness. In this regard, the results in \cref{sect:Hierarchy} showcase that instead of choosing a fixed approximation space from the beginning, we can gradually build a suitable approximation space by iteratively refining the space until it reaches the desired accuracy.
	But first, we discuss some algebraic prerequisites and give a short introduction to consistent interval-\AFT in \cref{sect:Prel}.
	
	
	\section{Preliminaries}\label{sect:Prel}
	Assume for the remainder of this section that $\CS = \langle \cpo, \leq \rangle$ is a \emph{partially ordered set (poset)}, \ie a set $\cpo$ equipped with a partial order $\leq$ (a reflexive, anti-symmetric and transitive relation). Let $S \subseteq \CS$, then $x \in \CS$ is an \emph{upper bound}, respectively a \emph{lower bound}, of $S$ if for every $s \in S$, $s \leq x$, respectively $x \leq s$. We say $x \in \CS$ is the \emph{least upper bound} of $S$, denoted by $\lub(S)$, if it is the least of all upper bounds of $S$, \ie $x$ is an upper bound of $S$ and $x \leq u$ for each upper bound $u$ of $S$. 
	Analogously, $x \in \CS$ is the \emph{greatest lower bound} of $S$, denoted by $\glb(S)$ if it is the greatest of all lower bounds of $S$, \ie $x$ is a lower bound of $S$ and for every lower bound $l \in \CS$, $l \leq x$. Note that the least upper bound or greatest lower bound of $S$ might not exist. The set of minimal elements of $S$, denoted by $\Min{S}$ is defined as $\{x \in S \mid \neg \exists y \in S: y < x\}$. Similarly, the set of maximal elements of $S$, denoted by $\Max{S}$ is defined as $\{x \in S \mid \neg \exists y \in S: x < y\}$. We say $S$ is a \emph{chain} if $\leq$ is total for $S$, \ie if for each two distinct elements $x, y \in S$, either $x \leq y$ or $y \leq x$. 
	In contrast, $S$ is an \emph{anti-chain} if all distinct elements in $S$ are incomparable, \ie if for each two elements $x, y \in S$ with $x\neq y$, $x \not \leq y$. A set $S \subseteq \CS$ is \emph{convex} if $\forall x, z \in S, \forall y \in \CS: x \leq y \leq z$ implies that $y \in S$. The set $S \subseteq \CS$ is \emph{closed} if it contains all its limit points, \ie if for every non-empty chain $R \subseteq S$, $\glb(R)$ and $\lub(R)$ exist and are in $S$. 
	The lower closure of a set $S \subseteq \CS$ is defined as 
	$\lc{S} = \{x \in \CS \mid \exists y \in S: x \leq y\}$. Similarly, the upper closure of a set $S \subseteq \CS$ is given by $\uc{S} = \{x \in \CS \mid \exists y \in S: y \leq x\}$. Then the lower and upper closures of an element $x \in \CS$, denoted by $\lc{x}$ and $\uc{x}$ respectively, are the lower and upper closures of the singleton $\{x\}$. Let $\imcons: \CS \to \CS$ be an operator on $\CS$. For any $x \in \CS$, we say $x$ is a fixpoint of $\imcons$ if $\imcons(x) = x$. Moreover, $x$ is a pre-fixpoint, respectively post-fixpoint, if $\imcons(x) \leq x$, respectively $x \leq \imcons(x)$. We call $x$ the least, respectively greatest, fixpoint of $\imcons$ if for every fixpoint $y$ of $\imcons$, $x \leq y$, respectively $y \leq x$. We denote the least fixpoint by $\lfp(\imcons)$ and the greatest fixpoint by $\gfp(\imcons)$, if they exist. The least and greatest pre- and post-fixpoints are defined analogously.
	
	\begin{definition}[Chain-complete partial order (cpo)]
		$\CS$ is \emph{a chain-complete partial order (cpo)} if every chain of $\CS$ has a least upper bound. 
	\end{definition}
	Note that every cpo has a least element $\bot$, which is the least upper bound of $\emptyset$. 	A cpo $\CS$ is \emph{bounded-complete} if every non-empty subset $S \subseteq \CS$ has a greatest lower bound. If $\CS$ is a bounded complete-cpo and $S \subseteq \CS$ has an upper bound, then it has a least upper bound. Moreover, if the empty set has a greatest lower bound as well, we say the cpo $\CS$ is a \emph{complete lattice}. 
	
%
%
	
	\begin{definition}[Monotone induction]\label{def:MonInduction}
		Let $\imcons$ be a monotone operator on a cpo $\CS$ then a \emph{monotone induction} is an increasing sequence $(x_i)_{i\leq\beta}$ satisfying 
		\begin{compactitem}
			\item $x_i \leq x_{i+1}\leq \imcons(x_i)$ , for successor ordinals $i+1\leq \beta$,
			\item $x_\lambda=\lub(\{x_i\mid i<\lambda\})$, for limit ordinals $\lambda\leq \beta$  (in particular, $x_0=\bot$).
		\end{compactitem}
	\end{definition}
	A monotone induction is terminal if its limit can not be refined further, i.e., if the limit is a prefixpoint of $\imcons$. 
	It has been shown that all terminal monotone inductions converge to the same limit \cite[Corollary 3.7]{ai/BogaertsVD18}. 
	\begin{theorem}[Knaster-Tarski theorem \cite{Tarski}]
		Given a cpo $\CS = \langle \cpo, \leq \rangle$, if $\imcons$ is a monotone operator on $\CS$, then $\imcons$ has a least fixpoint. In particular, the least fixpoint of $\imcons$ is the least pre-fixpoint of $\imcons$ and can be constructed as the limit of a terminal monotone induction of $\imcons$.
	\end{theorem}
	
	The Knaster-Tarski theorem provides a constructive way to obtain the least fixpoint of a monotone operator, \AFT generalizes this theorem to non-monotonic operators by using approximations. Given that $\CS$ is a complete lattice, we can construct the bilattice $\AbS = \langle \cpo^2, \leqp \rangle$
	, where for every $x_1, x_2, y_1, y_2 \in \CS$, $(x_1, x_2) \leqp (y_1, y_2)$ iff $x_1 \leq y_1$ and $y_2 \leq x_2$. It is easy to see that $\leqp$ is a partial order. In natural language we call this order the \emph{precision order}. Similarly, we define the \emph{truth order} $\leqt$ such that for every $x_1, x_2, y_1, y_2 \in \CS$, $(x_1, x_2) \leqt (y_1, y_2)$ if $x_1 \leq y_1$ and $x_2 \leq y_2$. Since we focus on consistent \AFT \cite{DeneckerMT04}, which is in general more natural and provides sufficient expressiveness, we use the consistent subspace\footnote{To limit the the number of symbols, we will in often use the same symbol for an order projected on a subspace, here $\leqp$ is used as a binary relation on the consistent subspace as well as the bigger bilattice.} $\AbS^c = \langle \cpo^c, \leqp \rangle$, where $\cpo^c = \{(x_1, x_2) \in \cpo^2 \mid x_1 \leq x_2\}$.
	
	\begin{definition}[Approximating Operator (Approximator)]
		An operator $\aop: \AbS^c \to \AbS^c$ is an \emph{approximating operator} if $\aop$ is $\leqp$-monotone. We denote the set of approximating operators on  $\AbS^c$ by $\appr[{\AbS^c}]$. We will often use the shorted term \emph{approximator} instead of approximating operator.
	\end{definition}
	
	We call any $(x_1, x_2) \in \AbS^c$ an \emph{approximant}, and we say that an approximant $(x_1, x_2)$ approximates an element $y \in \CS$ if $x_1 \leq y \leq x_2$. An approximant $(x_1, x_2) \in \AbS^c$ is exact if it approximates exactly one element, \ie if $x_1 = x_2$. An approximator $\aop$ approximates an operator $\imcons: \CS \to \CS$ if for every $(x_1, x_2) \in \AbS^c$ and $y \in \CS$, $\imcons(y)$ is approximated by $\aop(x_1, x_2)$ whenever $y$ is approximated by $(x_1, x_2)$. An approximator $\aop[1]$ is \emph{less precise} than an approximator $\aop[2]$, denoted by $\aop[1] \leqp \aop[2]$, if for every approximant $(x_1, x_2)$, $\aop[1](x_1, x_2) \leqp \aop[2](x_1, x_2)$. The most precise approximator $\ult[\imcons]$ which approximates an operator $\imcons$ is called the \emph{ultimate approximator}\footnote{When the operator is clear from the context, we will drop the subscript-notation.} of $\imcons$ and is given by $\ult[\imcons](x_1, x_2) = (\glb_{\CS}\{\imcons(y) \mid y \in [x_1, x_2]\}, \lub_{\CS}\{\imcons(y) \mid y \in [x_1, x_2]\})$, where $[x_1, x_2] = \{y \in \CS \mid x_1 \leq y \leq x_2\}$.
	
	\begin{definition}[Stable Revision Operator]
		Let $\aop$ be an approximator, then the \emph{stable revision operator} is given by $\sto: \AbS^c \to \AbS^c$ which maps any $(x_1, x_2) \in \AbS^c$ to $(\lfp(\aoplx{x}), \lfp(\aopux{x}))$, where $\aoplx{x}(y_1)$ is the projection of $\aop(y_1, x_2)$ on its first component and $\aopux{x}(y_2)$ is the projection of $\aop(x_1, y_2)$ on its second component.
	\end{definition}
	
	Given an approximator $\aop$, we say an approximant $(x_1, x_2)$ is $\aop$-reliable if $(x_1, x_2) \leqp \aop(x_1, x_2)$. Moreover, $(x_1, x_2)$ is $\aop$-prudent if $x_1 \leq \lfp(\aoplx{x})$. The following theorem summarizes an important result for the subspace of $\aop$-reliable and $\aop$-prudent elements of $\AbS^c$.
	
	\begin{theorem}[\cite{DeneckerMT04} Theorem 3.11]
		Let $\AbS^c = \langle \cpo^c, \leqp \rangle$ be a consistent approximation space with approximator $\aop$. The set of $\aop$-reliable and $\aop$-prudent elements of $\AbS^c$ is a cpo under the precision order $\leqp$ with least element $(\bot, \top)$. The stable revision operator is a well-defined, increasing and monotone operator in this poset.
	\end{theorem}
	
	Since $\aop$ is $\leqp$-monotone on $\AbS^c$, it has a least (precise) fixpoint, this is the \emph{Kripke-Kleene fixpoint} of $\aop$, denoted by $\KK(\aop)$. The set of \emph{supported fixpoints} of $\aop$, denoted by $\SUP(\aop)$, is defined as $\{x \in \CS \mid \aop(x, x) = (x, x)\}$, \ie the exact fixpoints of $\aop$. Similarly, $\sto$ is $\leqp$-monotone on the set of $\aop$-reliable and -prudent elements of $\AbS^c$. Therefore, it has a least (precise) fixpoint in that space, \ie the \emph{Well-founded fixpoint} of $\aop$, denoted by $\WF(\aop)$. The set of \emph{$\aop$-stable fixpoints}, denoted by $\ST(\aop)$, is defined as $\{x \in \CS \mid \sto(x, x) = (x, x)\}$, \ie the exact fixpoints of $\sto$. In many case, e.g., in the case of logic programming, these fixpoints correspond to the eponymous semantics. In other words, for any non-monotonic reasoning formalism we can determine the Kripke-Kleene, supported, Well-founded and stable semantics for free once we have defined a suitable approximator. Usually this approximator naturally follows from the application. 
	
	\section{Approximations with Multiple Maximal Elements}\label{sect:Flowers}
	While standard \AFT (interval-\AFT) already provides sufficient expressiveness for many non-monotonic reasoning formalisms, \cref{ex:Hanne,ex:Paper} showcase the limitations of this theory. In this section we examine a space of approximations sufficiently refined to solve the issues for both examples, \ie the space of \emph{flowers}. Within this space, we identify and analyze a set of requirements and promising properties which will serve as inspiration for the definition of an approximation space in \cref{sect:Theory}. Note, however, that while the space serves well as an inspiring and motivating example, it is definitely \emph{not} the only instance of an approximation space covered by our framework. 
	
	Let us first consider \cref{ex:Hanne} in a bit more detail. The main semantic objects of AEL are \emph{belief states}; these are sets of interpretations that the agent in question deems possible. The agent \emph{knows} a formula if it is true in all the interpretations in its belief state. For instance, in the belief state that consists of all the interpretations where $p$ is true, $Kp$ holds, as will $K(q\lor \lnot q)$, but $K(q\land p)$ does not hold. 
	The belief states are ordered by the superset order, hence the greatest belief state $\top$ is the empty set where no interpretation is deemed possible, \ie the inconsistent belief state. On the other hand, the least belief state $\bot$ is the set of all interpretations. Note that `greatest' and `least' refer here to the amount of information or knowledge in the belief state. 
	Given a belief state $\mathcal{X}$, we can evaluate all formulas of the form $K\varphi$. 
	Now it is easy to derive a suitable operator for \cref{ex:Hanne}, namely the operator $\imcons$ which maps any belief state $\mathcal{X}$ to the
	set $\mathcal{Y}$ of all interpretations $Y$ such that \emph{(1)} $q$ is true in $Y$ iff $K p$ is false in $\mathcal{X}$ and \emph{(2)} $r$ is true in $Y$ iff $K q$ is false in $\mathcal{X}$. 
%
%
It is easy to verify that the ultimate approximator in interval-\AFT for this operator derives as fixpoints $\KK(\ult) = \WF(\ult) = (\bot, \top)$. 
	Our intuitions for \cref{ex:Hanne} tell us that since there is no direct information about $p$, $p$ is not known. From this we then derive all other information. In our minds we lower the upper bound for our belief sets, \ie we only consider belief sets where $p$ is not known. 
	The problem is that among those there are incompatible pairs of belief states, \ie their intersection is empty, e.g., any belief state where we know $q$ is incompatible with a belief state where we know $\neg q$. 
	The only approximant that approximates all these belief states is $(\bot, \top)$. In other words, intervals are not sufficiently precise to represent the upper bound we had in our mind, therefore we lose all the acquired information. On the other hand, by introducing approximants with an upper bound consisting of multiple maximal elements, we would be able to perfectly represent the set of belief states where $p$ is not known. This leads us to the idea of \emph{flowers}.   
	
	\begin{definition}[Lower-bounded Convex Set (Flower)]\label{def:LowerBounded}
		Given a bounded-complete cpo $\CS = \langle \cpo, \leq \rangle$, the set $\mathfrak{X} \subseteq \CS$ is \emph{lower-bounded convex} if $\mathfrak{X}$ is closed and convex, and 
		$\glb(\mathfrak{X}) \in \mathfrak{X}$. We say $\mathfrak{X}$ is a \emph{flower}.
		We use $\flower$ to denote the set of flowers over the bounded-complete cpo $\CS$.
	\end{definition}
	Informally, flowers are unions of intervals that have a common greatest lower bound. The set of belief states where $p$ is not known is a flower. We say a flower $\mathfrak{X}$ approximates an element $x \in \cpo$ if $x \in \mathfrak{X}$. Then it is sensible to define a precision-order on $\flower$, \ie given two flowers $\mathfrak{X}, \mathfrak{Y}\in \flower$, then $\mathfrak{X} \leqp \mathfrak{Y}$ if $\mathfrak{X} \supseteq \mathfrak{Y}$. It can be shown that $\langle \flower, \leqp \rangle$ is a cpo\footnote{All proofs are in the supplementary material due to space limitations.}.
	
	Suppose we are given a generalized approximator on the set of flowers over $\CS$ and we want to extend AFT to work with this. 
	The fact that this set is a cpo is essential if we want to use the Knaster-Tarski theorem to construct the Kripke-Kleene fixpoint. 
	\begin{require}
		A generalized approximator must operate on a cpo.
	\end{require}
	
	A crucial definition in AFT is the stable revision operator. 
	For interval-\AFT this operator reasons on a `lower' and `upper' element (corresponding to the greatest lower bound and least upper bound of the interval). Hence, in order to stay close to its original definition, we require a decomposition of our approximations.
	\begin{require}
		An approximation must be decomposable into an approximation lower bound (\alb) and an approximation upper bound (\aub).
	\end{require}
	
	Naturally a flower $\mathfrak{X}$ has as an \alb $\low[\mathfrak{X}] = \glb(\mathfrak{X})$. On the other hand, we can represent the \aub $\up[\mathfrak{X}]$ by $\Max{\mathfrak{X}}$ since for every $x \in \mathfrak{X}$ there will exist at least one $m \in \Max{\mathfrak{X}}$ such that $x \leq m$. This brings about the decomposition spaces.
	
	\begin{definition}[Decomposition Spaces of Flowers]
		Given a bounded-complete cpo $\langle \cpo, \leq \rangle$, we define 
		\begin{compactitem}
			\item $\lowlat[f] = \{\low[\mathfrak{X}] \mid \mathfrak{X} \in \flower\} = \cpo$.
			\item $\uplat[f] = \{\up[\mathfrak{X}] \mid \mathfrak{X} \in \flower\} =\{\emptyset \subsetneq S \subseteq \cpo \mid \lc{S} \text{ is closed } \land S \text{ is an anti-chain}\}$.
		\end{compactitem}
	\end{definition}
	
	Moreover, the stable operator is defined as the combination of two least fixpoints, this suggests we need an order on our decomposition spaces. Since we think of such elements as bounds of our flowers, it is natural to expect that increasing the \alb or decreasing the \aub should correspond to a more precise flower. Moreover, since we are extending consistent \AFT, we also require that the \alb of a flower is smaller than its \aub.
	
	\begin{require}
		The (union of the) decomposition spaces should be equipped with an order $\leqA$ such that for every pair of approximants $\mathfrak{X}, \mathfrak{Y}$, $\mathfrak{X} \leqp \mathfrak{Y}$ implies $\low[\mathfrak{X}] \leqA \low[\mathfrak{Y}] \leqA \up[\mathfrak{Y}] \leqA \up[\mathfrak{X}]$.
	\end{require}
	
	\begin{definition}
		Given a bounded complete cpo $\langle \cpo, \leq \rangle$, we define the (flower) composition order $\leqf$ s.t. for every $b_1, b_2 \in \lowlat[f] \cup \uplat[f]$, $b_1 \leqf b_2$ if $\lc{b_1} \subseteq \lc{b_2}$, and $b_1 \in \lowlat$ or $b_2 \in \uplat$.
	\end{definition}
	 
	If for $l_1, l_2 \in \lowlat[f]$, $\l_1 \leqf l_2$, we say $l_1$ is a less precise \alb than $l_2$. On the other hand, if for $U_1, U_2 \in \uplat[f]$, $U_1 \leqf U_2$, we say $U_1$ is a more precise \aub than $U_2$. 
		
	\begin{figure}
		\centering
		\begin{tikzpicture}[node distance=0.2cm]
			\node (a) at (0, 1) {$a$};
			\node (b) at (2, 1) {$b$};
			\node (bot) at (1, 0) {$\bot$};				
			\path[draw,-]
			(a) edge node {} (bot)
			(b) edge node {} (bot);		
		\end{tikzpicture}
		\caption{A simple bounded-complete cpo. For this cpo, the set of non-empty flowers $\flower$ consists of $\{\bot\}, \{a\}, \{b\}, \{\bot, a\}, \{\bot, b\}$ and $\{\bot, a, b\}$. Then $\lowlat[f]$ consists of $\bot, a$ and $b$ and $\uplat[f]$ consists of $\{\bot\}, \{a\}, \{b\}$ and $\{a, b\}$.}
		\label{fig:BoundedCPO}
		\vspace{-0.4cm}
	\end{figure}
	
	Once both fixpoints are determined, the stable revision operator needs to (re)combine the information about the \alb and the \aub back into a flower. Clearly, for every $l \in \lowlat[f]$ and $U \in \uplat[f]$ such that $l \leqf U$, we can recompose them into a flower by taking all elements $x \in \cpo$ such that $l \leq x$ (\ie $x$ is greater than the \alb) and $x \in \lc{U}$ (\ie $x$ lies under the \aub), \ie the recomposition $(l, U)$ is given by $\uc{l} \cap \lc{U}$. However, it is important to see that given an arbitrary $l \in \lowlat[f]$ and $U \in \uplat[f]$ such that $l \leqf U$, decomposing the recomposition $(l, U)$ will not always result in the same \alb and \aub. E.g., consider the cpo in \cref{fig:BoundedCPO}, clearly $\{\bot, a, b\} \leqp \{a\}$, therefore $\bot \leqf a \leqf \{a\} \leqf \{a, b\}$. The recomposition $(a, \{a, b\})$ corresponds to the flower $\{a\}$ since $b$ is not greater than the lower bound $a$. Therefore, $\up[\{a\}] = \{a\} \neq \{a, b\}$. Intuitively, recomposing a lower bound and upper bound combines the information on the lower bound and the upper bound to derive additional knowledge about these bounds. It is then natural to expect that such recomposition results in a more precise lower and upper bound after decomposition. On the other hand, recomposing the decomposition of a flower $\mathfrak{X}$ always obtains the exact same flower, \ie $\mathfrak{X} = (\low[\mathfrak{X}], \up[\mathfrak{X}]) = (\glb(\mathfrak{X}), \Max{\mathfrak{X}})$. 
	
	\begin{require}
		We need a recombination function capable of recombining an \alb with an \aub into an approximant, provided they are compatible. Such a recombination might improve the \alb and \aub, but it should never lose information.
	\end{require}
	
	Finally, part of the strength of \AFT lies in the fact that we can essentially use the Knaster-Tarski theorem to derive the well-founded model. For this the stable revision operator needs to be monotonic and it should operate on a cpo.
	\begin{require}
		The stable revision operator must be a monotonic operator on a cpo.
	\end{require}
	
	In this regard we identify four interesting characteristics exhibited by the recomposition and decomposition operations for flowers.
	\begin{prop}\label{prop:FlowersChainILP}
		If we assume a set of maximal elements $U \in \uplat[f]$ and a chain $S \subseteq \lowlat[f]$ such that for every $l \in S$ there exists an $u \in U$ such that $l \leq u$, then there must exist an $u' \in U$ such that $\lub_{\lowlat[f]}(S) = \lub_{\CS}(S) \leq u'$, therefore $\lub_{\lowlat[f]}(S) \leqf U$
	\end{prop}
	This property follows from the fact that $\lc{U}$ is closed. Informally, this property ensures that when we restrict our original bounded-complete cpo to everything below below certain elements $U$, we retain a bounded-complete cpo. Therefore, as long as our operator is internal on this restricted cpo, which corresponds to $\aop$-reliability, we can essentially zoom in on our approximation space to the part pertaining only to this restricted cpo and treat it as though it is the whole approximation space. Note that since $\lowlat[f] = \cpo$ is only a bounded complete cpo and not a complete lattice, not every set $S \subseteq \lowlat[f]$ will have a least upper bound. However, if $S$ has an upper bound, then it will also have a least upper bound. This brings us to the second interesting property. 
	
	\begin{prop}\label{prop:FlowersWeakILP}
		Let $\mathfrak{X} \in \flower$ and $l \in \lowlat[f]$. If $l \leqf \Max{\mathfrak{X}}$, then $\lub_{\lowlat[f]}\{l, \glb(\mathfrak{X})\}$ exists and $\lub_{\lowlat[f]}\{l, \glb(\mathfrak{X})\} \leqf \Max{\mathfrak{X}}$.
	\end{prop}
	This follows from the fact that for every $m \in \Max{\mathfrak{X}}$, $\glb(\mathfrak{X}) \leq m$. Therefore, since $l \leqf \Max{\mathfrak{X}}$, there must exist an $m \in \Max{\mathfrak{X}}$ such that $m$ is an upper bound of $\{l, \glb(\mathfrak{X})\}$. This second property ensures that if we have current knowledge about the bounds of an approximation and we derive new knowledge about the \alb that respects the current knowledge about our \aub, than this new knowledge is compatible with our current knowledge. 
	
	\begin{prop}\label{prop:FlowersAbstractILP}
		Let $S \subseteq \flower$ be a set of flowers such that they all have the same \aub $U \in \uplat[f]$, \ie for every $\mathfrak{X} \in S$, $\up[\mathfrak{X}] = \Max{\mathfrak{X}} = U$. Then $\lub_{\lowlat}\{\glb(\mathfrak{X}) \mid \mathfrak{X} \in S\}$ exists, and $\low[(\lub_{\flower}(S))] = \lub_{\lowlat}\{\glb(\mathfrak{X})\mid \mathfrak{X} \in S\}$ and $\up[(\lub_{\flower}(S))] = U$.
	\end{prop}
	Clearly, $\mathfrak{X} \in S$ entails that $\glb(\mathfrak{X}) \leq u$ for every $u \in U$, therefore, every $u \in U$ is an upper bound of $\{\glb(\mathfrak{X}) \mid \mathfrak{X} \in S\}$. Then the proposition above trivially follows. This property formalizes two natural observations. On the one hand, it suggests that the knowledge derived from a set of \albs can only be mutually incompatible when their recomposition with the \aub $U$ derives additional information about the \aub, which might be contradictory. On the other hand it guarantees that combining the knowledge captured by the \albs will only improve our knowledge on the \alb and not on the \aub, which would be surprising since none of the \albs provides additional information about the \aub on its own. This property is required to ensure monotonicity, \ie more precise flowers acquire more knowledge. 
	
	\begin{prop}\label{prop:FlowersIGP}
		Let $l \in \lowlat[f]$ and $S \subseteq \uplat[f]$. If in every $U \in S$ there exists an $u \in U$ such that $l \leq u$, then $l \leqf \glb_{\uplat[f]}(S)$.
	\end{prop}
	It is easy to verify that this property entails analogous properties for the greatest lower bound as Propositions \ref{prop:FlowersChainILP} to \ref{prop:FlowersAbstractILP}. However, since we need a cpo, and thus a least precise \aub, $\uplat[f]$ requires a greatest element, in other words, $\uplat[f]$ needs to be a complete lattice. Therefore, in addition to the aforementioned characteristics, this final property ensures that increasing the \alb of our bounded-complete cpo again results in a space of \aubs that is a complete lattice. Note, however, that the analogous version of this last property does not hold for the least upper bound. For one, the $\lub_{\lowlat[f]}(S)$ does not always exist, but even if it exists, it might still not hold. For example, consider \cref{fig:BoundedCPO} where we add a greatest element $\top$ to obtain a complete lattice. Then $a \leqf \{a, b\}$ and $b \leqf \{a, b\}$, however, $\lub_{\lowlat[f]}\{a, b\} = \top \not \leqf \{a, b\}$.
	
	\section{Generalization of consistent \AFT}\label{sect:Theory}
	 In this section we provide a general notion of an approximation framework and approximation space heavily inspired by the properties of flowers discussed above. The main results of interval-\AFT remain valid even after this generalization.
	 
	 First, we require that our approximants in the approximation space can be decomposed into a \alb and an \aub. Whenever the \alb and \aub are compatible, we want to be able to recompose them back into an approximant. Moreover, more precise \albs and \aubs should lead to more precise approximants. This is summarized in the following definition of a composition poset.
	 \begin{definition}[Composition poset]\label{def:Composition}
	 	Let $\CS = \langle \lowlat \cup \uplat, \leqA \rangle$ be a poset. A poset $\langle \AbS, \leqp \rangle$ is a \emph{composition poset} of $\CS$ if there exist decomposition-functions $\low: \AbS \to \lowlat$ and $\up: \AbS \to \uplat$, and partial recomposition-function $(., .): \lowlat \times \uplat \to \AbS$ such that for every $l, l_1, l_2 \in \lowlat$ and $u, u_1, u_2 \in \uplat$:
	 	\begin{compactitem}
	 		\item $(l, u)$ is defined if $l \leqA u$.
	 		\item $l \leqA \low[(l, u)]$ and $\up[(l, u)] \leqA u$.
	 		\item if $l_1 \leqA l_2$, then $(l_1, u) \leqp (l_2, u)$.
	 		\item if $u_1 \leqA u_2$, then $(l, u_2) \leqp (l, u_1)$.
	 		\item for every $\mathfrak{X} \in \AbS$, $\mathfrak{X} = (\low[\mathfrak{X}], \up[\mathfrak{X}])$.
	 	\end{compactitem}
	 	In natural language, we will refer to $\leqp$ as the precision order, and to $\low[\mathfrak{X}]$ and $\up[\mathfrak{X}]$ as the \alb and \aub of $\mathfrak{X}$, respectively.
	 \end{definition} 
	 
	 With this, we can define the notion of an approximation framework,\footnote{Note that interval-\AFT, as well as the generalization of \AFT proposed by \citet{tplp/CharalambidisRS18}, are special instances of this general notion, but neither captures flowers.} explicitly linking the composition poset and the spaces of \albs ($\lowlat$) and \aubs ($\uplat$). An attentive reader will notice the strong similarities between properties 2-5 in the following definition and Properties 1-4 for flowers described above.
	 
	 \begin{definition}[Approximation framework]\label{def:ApSpaceAlt}
	 	Assume that $\lowlat$ and $\uplat$ are sets and that $\leqA$ is a partial order over $\lowlat \cup \uplat$ such that $\langle \lowlat \cup \uplat, \leqA \rangle$ has a least element $\bot \in \lowlat$ and a greatest element $\top \in \uplat$, $\langle \lowlat, \leqA \rangle$ is a bounded-complete cpo and $\langle \uplat, \leqA \rangle$ is a complete lattice. An \emph{approximation framework} $\AF$ is a tuple $\langle \lowlat, \uplat, \AbS, \leqA \rangle$ such that 
	 	\begin{enumerate}
	 		\item $\langle \AbS, \leqp \rangle$ is a composition poset of $\langle \lowlat \cup \uplat, \leqA \rangle$.
	 		\item it satisfies the \emph{\ChainILP}: Let $u \in \uplat$ and $S \subseteq \lowlat$ be a chain. If for every $l \in S$, $l \leqA u$, then $\lub_{\lowlat}(S) \leqA u$.
	 		\item it satisfies the \emph{\WeakILP}: Let $\mathfrak{X} \in \AbS$ and $l \in \lowlat$. If $l \leqA \up[\mathfrak{X}]$, then $\lub_{\lowlat}(\low[\mathfrak{X}], l)$ exists and $\lub_{\lowlat}(\low[\mathfrak{X}], l) \leqA \up[\mathfrak{X}]$.
	 		\item it satisfies the \emph{\AbstractILP}: If $S \subseteq \AbS$ is a set of elements such that $\up[\mathfrak{X}] =u$ for every $\mathfrak{X}\in S$, then $\lub_{\AbS}(S)$ exists, and $\up[(\lub_{\AbS}(S))] = u$ and $\low[(\lub_{\AbS}(S))] = \lub_{\lowlat}\{\low[\mathfrak{X}] \mid \mathfrak{X} \in S\}$.
	 		\item it satisfies the \emph{\IGP}: Let $l \in \lowlat$ and $S \subseteq \uplat$. If for every $u \in S$, $l \leqA u$, then $l \leqA \glb_{\uplat}(S)$.	
	 	\end{enumerate} 
	 \end{definition}
	 
	 Now it is trivial to define the space of approximants, \ie the approximation space, given an approximation framework.
	 \begin{definition}[Approximation Space and Approximants]
	 	Let $\AF = \langle \lowlat, \uplat, \AbS, \leqA \rangle$ be an approximation framework. We call $\langle \AbS, \leqp \rangle$ the approximation space, and any $\mathfrak{X} \in \langle \AbS, \leqp \rangle$ an \emph{approximant}. 
	 \end{definition}
	An approximant is \emph{exact} if it is $\leqp$-maximal or if it approximates exactly one $\leqp$-maximal approximant.
	It should be clear that the space of flowers as described above is an instance of an approximation space where the singletons are the exact approximants, as expected.
	
	Our goal was to build a conservative extension of interval-\AFT. Therefore, the remaining definitions in this section are natural generalizations of their counterparts in interval-\AFT. As such, from this point forward, the challenge lies in proving that these generalizations are well-defined and that the main theorems still hold.  
	
	\begin{definition}[Approximating Operator (Approximator)]\label{def:AppOp}
		Let $\AF = \langle \lowlat, \uplat, \AbS, \leqA \rangle$ be an approximation framework. 
		We say that $\aop: \AbS \to \AbS$ is an \emph{approximating operator} of \emph{approximator} if $\aop$ is $\leqp$-monotone. We use $\appr$ to denote the set of approximators for $\AF$.
	\end{definition}
	
	\begin{definition}[Stable Revision Operator]
		Let $\AF = \langle \lowlat, \uplat, \AbS, \leqA \rangle$ be an approximation framework with approximator $\aop$. 
		The \emph{stable revision operator} $\sto[\aop]: \AbS \to \AbS$ maps any approximant $\mathfrak{X} \in \AbS$ to $(\stl[\aop](\mathfrak{X}), \stu[\aop](\mathfrak{X}))$, where 
		\begin{gather*}
			\stl[\aop](\mathfrak{X}) = \lfp(\aopllx{\mathfrak{X}})\\
			\stu[\aop](\mathfrak{X}) = \lfp(\aopuux{\mathfrak{X}})
		\end{gather*}
		where $\aopllx{\mathfrak{X}}: \lowlat \to \lowlat: l \mapsto \low[(\aop(l, {\up[\mathfrak{X}]}))]$ and similarly, $\aopuux{\mathfrak{X}}: \uplat \to \uplat: u \mapsto \up[(\aop({\low[\mathfrak{X}]}, u))]$.
	\end{definition}
	
	Similar to interval-\AFT, we say an approximant $\mathfrak{X}$ is \emph{$\aop$-reliable}, if $\mathfrak{X} \leqp \aop(\mathfrak{X})$. Moreover, $\mathfrak{X}$ is \emph{$\aop$-prudent} if $\low[\mathfrak{X}] \leqA \stl[\aop](\mathfrak{X})$. It can be shown that if $\mathfrak{X}$ is $\aop$-reliable, then $\aopllx{\mathfrak{X}}$ and $\aopuux{\mathfrak{X}}$ are monotone operators on a cpo, therefore by the Knaster-Tarski theorem they have a least fixpoint. Furthermore, the subspace of $\aop$-reliable and $\aop$-prudent approximants is a cpo and the stable revision operator is monotone on this cpo. This yields the main theorem of \AFT.
	
	\begin{theorem}
		Let $\AF = \langle \lowlat, \uplat, \AbS, \leqA \rangle$ be an approximation framework with approximator $\aop$. The set of $\aop$-reliable and $\aop$-prudent elements of $\AbS$ is a cpo under the precision order $\leqp$ with least element $(\bot, \top)$. The stable revision operator is a well-defined, increasing and monotone operator in this poset.
	\end{theorem}
	
	Now we are able to define the different types of fixpoints analogous to how they were defined in interval-\AFT. The Kripke-Kleene and Well-founded fixpoints are the least fixpoints of the approximating and stable revision operator, respectively. On the other hand, the sets of supported models and stable models are defined as the set of exact fixpoints of the approximator and the stable revision operator, respectively. 
	To constructively derive the Kripke-Kleene fixpoint of an approximator it suffices to look at monotone inductions of $\aop$. Similarly, for the Well-founded fixpoint, \citet{lpnmr/DeneckerV07} introduced a similar natural construction process for interval-\AFT, using the notion of \emph{refinements}. This work also naturally generalizes to the relaxed definition of approximation spaces. 
	\begin{definition}[Refinement]
		Let $\langle \AbS, \leqp \rangle$ be an approximation space with s.t. $\aop \in \appr$. 
		
		$\mathfrak{Y} \in \AbS$ is an \emph{application refinement} of $\mathfrak{X} \in \AbS$ if $\mathfrak{X} \leqp \mathfrak{Y} \leqp \aop(\mathfrak{X})$
		
		$\mathfrak{Y} \in \AbS$ is a \emph{grounding refinement} of $\mathfrak{X} \in \AbS$ if there exists an $u \in \uplat$ such that $\low[\mathfrak{X}] \leqA u \leqA \up[\mathfrak{X}] \text{ and } \mathfrak{Y} = (\low[\mathfrak{X}], u) \text{ and } \mathfrak{Y} \leqp \aop(\mathfrak{Y})$
	\end{definition}
	
	Informally, an application refinement corresponds to applying a possibly less precise version of the approximator on our approximant, while a grounding refinement corresponds to deleting an \emph{unfounded set} of elements from our approximant. For example, assuming we are using the ultimate approximator, it is easy to verify that when we use the space of flowers for \cref{ex:Hanne}, imposing the intuitive upper bound where $p$ is not known corresponds to a grounding refinement. Moreover, the other two intuitive steps are essentially application refinements.
	
	\begin{definition}[Well-founded induction]\label{def:WFInduction}
		Let $\AbS = \langle \cpo^c, \leqp \rangle$ be a consistent approximation space with approximator $\aop$. A \emph{well-founded induction} of $\aop$ is an increasing sequence $(x^i_1, x^i_2)_{i \leq \beta}$ with $x^i_1, x^i_2 \in \CS$, satisfying 
		\begin{compactitem}
			\item $(x^{i+1}_1, x^{i+1}_2)$ is a refinement of $(x^i_1, x^i_2)$, for successor ordinals $i + 1 \leq \beta$,
			\item $(x^\lambda_1, x^\lambda_2) = \lub_{\AbS}\{(x^i_1, x^i_2) \mid i < \lambda\}$, for limit ordinals $\lambda \leq \beta$.
		\end{compactitem}
	\end{definition}
	
	Similar to the case of a monotone induction, a well-founded induction is terminal if its limit does not have a strictly more precise refinement and all terminal well-founded inductions converge to the same limit, \ie the well-founded model. From this, we can conclude that the ultimate semantics with flowers will indeed find the desired well-founded model for \cref{ex:Hanne}, based on the observations we made above. 
	
	\begin{theorem}\label{prop:WFIndConfluency}
		Let $\AF = \langle \lowlat, \uplat, \AbS, \leqA \rangle$ be an approximation framework with approximator $\aop$, the well-founded fixpoint of $\aop$ can be constructed as the limit of a terminal well-founded induction of $\aop$. 
	\end{theorem} 
	
	\section{Hierarchy of Approximation Spaces and Approximation Operators}\label{sect:Hierarchy}
	As is the case for interval-\AFT, we will use an approximator $\aop$ to approximate a (non-monotonic) operator $\imcons$ on the exact space $\CS = \langle \cpo, \leq \rangle$. However, contrary to interval-\AFT, the approximation space is not a fixed entity given an exact space. In fact there is a whole range of approximation spaces per exact space. This brings up the question of how to select an appropriate approximation space. This section shows that stronger or \emph{more precise} approximation spaces in general lead to semantics that are more precise. 
	For an approximation space $\langle \AbS, \leqp \rangle$ to approximate the exact space $\CS$ we first need to determine an `approximates'-relation between approximants and exact elements. In the case of intervals and flowers, we know that an approximant $\mathfrak{X}$ approximates an exact element $y \in \CS$ if $y \in \mathfrak{X}$. In general, given an approximant $\mathfrak{X} \in \AbS$ and an exact element $y \in \CS$, we will denote `$\mathfrak{X}$ approximates $y$' by $\mathfrak{X} \sim y$. Note that it is impossible to give a general formal definition for this relation, since this depends on the chosen approximation space. However, we hypothesize that for many natural approximation spaces this relation can be easily derived.
	
	\begin{definition}\label{def:Approximates}
		Let $\CS = \langle \cpo, \leq \rangle$ be a poset and let $\AF = \langle \lowlat, \uplat, \AbS, \leqA \rangle$ be an approximation framework. We say the approximation space $\langle \AbS, \leqp \rangle$ approximates $\CS$ if 
		\begin{compactitem}
			\item for every $\mathfrak{X}, \mathfrak{Y} \in \AbS$, $\mathfrak{X} \leqp \mathfrak{Y}$ implies that if for some $x \in \cpo$, $\mathfrak{Y} \sim x$, then $\mathfrak{X} \sim x$.
			\item for every $x, y \in \cpo$ and $l \in \lowlat$, $x \leq y$ implies that if $(l, \top) \sim x$, then $(l, \top) \sim y$.
			\item for every $x, y \in \cpo$ and $u \in \uplat$, $x \leq y$ implies that if $(\bot, u) \sim y$, then $(\bot, u) \sim x$.
			\item for every $\mathfrak{X} \in \AbS$, $\mathfrak{X}$ is exact iff there exists exactly one $x \in \cpo$ such that $\mathfrak{X} \sim x$.
		\end{compactitem}
	\end{definition}
	
 	For example, it is easily verified that if the exact space $\CS$ is a complete lattice, the sets of intervals and of flowers over $\CS$, ordered by the superset order, are approximation spaces of $\CS$. Informally, intervals are less precise than flowers, \ie flowers will be capable of approximating a set in the exact space more accurately than intervals. We have seen that intervals fail to solve \cref{ex:Hanne}. On the other hand, using the notion of refinements, we informally showed that flowers manage to find the intended well-founded model. Thus, flowers-\AFT seems to be stronger than interval-\AFT. On the other hand, the upper decomposition space of flowers is a lot larger than that of intervals. While a deep analysis of the complexity is beyond the scope of this paper, it is clear that in general, a larger upper decomposition space may have longer chains and will therefore result in an increase of required computational power to calculate the \aub for the stable revision operator. Hence, a good approximation space balances expressiveness with computational complexity. In this regard, we end the section with a proposal to iteratively construct a well-balanced approximation space for an application.
 	
 	\begin{definition}\label{def:ASPrecision}
 		Let $\AF[1] = \langle \lowlat[1], \uplat[1], \AbS[1], {\leqA}_1 \rangle$ and $\AF[2] = \langle \lowlat[2], \uplat[2], \AbS[2], {\leqA}_2 \rangle$ be approximation frameworks. We say $\langle \AbS[2], \leqp^2 \rangle$ is more precise than $\langle \AbS[1], \leqp^1 \rangle$, denoted by $\AbS[1] \leqp \AbS[2]$, if $\AbS[1] \subseteq \AbS[2]$, every exact approximant in $\AbS[1]$ is exact in $\AbS[2]$ and there exist a monotone function $\gal: \AbS[2] \to \AbS[1]$ such that 
 		\begin{equation*}
 			\forall \mathfrak{X}_1 \in \AbS[1], \mathfrak{X}_2 \in \AbS[2]: \mathfrak{X}_1 \leqp \mathfrak{X}_2 \text{ iff } \mathfrak{X}_1 \leqp \gal(\mathfrak{X}_2)
 		\end{equation*}
 	\end{definition}
 	
 	Intuitively, the monotone function $\gal$ is used to map an approximant $\mathfrak{X}_2$ in the more precise approximation space to the most precise approximant $\gal(\mathfrak{X}_2) \in \AbS[1]$ that approximates all exact elements $y \in \CS$ approximated by $\mathfrak{X}_2$. For example, in the case of flowers and intervals, it would map a flower $\mathfrak{X}$ to the interval $[\glb(\mathfrak{X}), \lub(\mathfrak{X})]$. 
 	
 	Naturally, one would assume that more precise approximation spaces provide more accurate semantics. However, this assertion is not entirely straightforward since the accuracy also strongly depends on the chosen approximator. Just as in interval-\AFT, we can have multiple approximators $\aop$ that approximate an operator $\imcons$ on the exact space. 
 	
 	\begin{definition}
 		Let $\imcons$ be an operator on a poset $\CS = \langle \cpo, \leq \rangle$ and let $\AF = \langle \lowlat, \uplat, \AbS, \leqA \rangle$ be an approximation framework with approximator $\aop$. We say $\aop$ approximates $\imcons$ if for every $\mathfrak{X} \in \AbS$ and $x \in \CS$, $\mathfrak{X} \sim x$ implies $\aop(\mathfrak{X}) \sim \imcons(x)$.
 	\end{definition}
 	
 	
 	\begin{definition}
 		Let $\AF = \langle \lowlat, \uplat, \AbS, \leqA \rangle$ be an approximation framework such that $\aop, B \in \appr$. We say $B$ is more precise than $\aop$, denoted by $\aop \leqp B$, if for every $\mathfrak{X} \in \AbS$, $\aop(\mathfrak{X}) \leqp B(\mathfrak{X})$.
 	\end{definition}
 	
	We adopt the term \emph{ultimate operator} from interval-\AFT to denote the most precise approximator $\ult$ of $\imcons$. 
	The level of precision of the chosen approximator influences the accuracy of the semantics as is formalized in the following theorem which is a generalization of Theorem 5.2 in the work of \citet{Denecker04}.
 	
 	\begin{theorem}\label{prop:LeastMorePrecise}
 		Let $\AF = \langle \lowlat, \uplat, \AbS, \leqA\rangle$ be an approximation framework with $\aop, B \in \appr$. If $\aop \leqp B$, then 
 		
 		\begin{minipage}{0.47\linewidth}
 			\begin{compactitem}
 				\item $\KK(\aop) \leqp \KK(B)$.
 				\item $\WF(\aop) \leqp \WF(B)$.
 			\end{compactitem}
 		\end{minipage}
 		\begin{minipage}{0.47\linewidth}
 			\begin{compactitem}
 				\item $\SUP(\aop) \subseteq \SUP(B)$.
				\item $\ST(\aop) \subseteq \ST(B)$.
 			\end{compactitem}
 		\end{minipage}
 	\end{theorem}
 	
 	While we cannot give a general statement about the relative accuracy of approximating spaces for any approximator, it is clear that the sets of approximators are strongly related. Intuitively, if $\AbS[1] \leqp \AbS[2]$, any approximator on 
 	$ \AbS[1]$ 	
can be transformed into an approximator on $\AbS[2]$ 
and vice versa. From \cref{def:ASPrecision}, we know there is a monotone function $\gal: \AbS[2] \to \AbS[1]$ that preserves the order. Therefore, if $\aop[1]$ is an approximator on $\AbS[1]$, \ie a $\leqp$-monotone operator on $\AbS[1]$, then the composition\footnote{Function composition is defined as usual, \ie given functions $f: D_2 \to D_3$ and $g: D_1 \to D_2$, the function $(f \circ g): D_1 \to D_3$ maps any $x \in D_1$ to $f(g(x))$.} $(\aop[1] \circ \gal)$ is a monotone operator on $\AbS[2]$ and thus an approximator on $\AbS[2]$. Similarly, if $\aop[2] \in \appr[{\AbS[2]}]$, then $(\gal \circ \aop[2])$ is an approximator on $\AbS[1]$. The following properties summarize how the semantics associated with such pairs of operators relate. 
 	 	
 	In the remaining part of the section we assume that $\langle \AbS[1], \leqp^1 \rangle$ and $\langle\AbS[2], \leqp^2 \rangle$ are approximation frameworks.
 	
 	\begin{theorem}\label{prop:FixpointPreserving}
 		Let $\AbS[1] \leqp \AbS[2]$. If $\aop[1] \in \appr[{\AbS[1]}]$ and $\aop[2] = (\aop[1] \circ \gal)$, then
 		\begin{compactitem}
 			\item $\mathfrak{X}$ is a fixpoint of $\aop[1]$ iff $\mathfrak{X}$ is a fixpoint of $\aop[2]$.
 			\item $\mathfrak{X}$ is a fixpoint of $\sto[{\aop[1]}]$ iff $\mathfrak{X}$ is a fixpoint of $\sto[{\aop[2]}]$.
 		\end{compactitem}
 		therefore $\KK(\aop[2]) = \KK(\aop[1])$ and $\WF(\aop[2]) = \WF(\aop[1])$
 	\end{theorem}
 	
 	\begin{theorem}
 		Let $\AbS[1] \leqp \AbS[2]$. If $\aop[2] \in \appr[{\AbS[2]}]$ and $\aop[1] = (\gal \circ \aop[2])$, then 
			\begin{compactitem}
				\item $\KK(\aop[1]) \leqp \KK(\aop[2])$.
				\item $\WF(\aop[1]) \leqp \WF(\aop[2])$.
				\item $\SUP(\aop[1]) \subseteq \SUP(\aop[2])$.
				\item $\ST(\aop[1]) \subseteq \ST(\aop[2])$.
			\end{compactitem}
 	\end{theorem}
 	
	
	\begin{theorem}                                                                             
		Let $\AbS[1] \leqp \AbS[2]$. Let $\ult[2] \in \appr[{\AbS[2]}]$ be the ultimate approximator of $\imcons$ for $\AbS[2]$, then $(\gal \circ \ult[2])$ is the ultimate approximator of $\imcons$ for $\AbS[1]$.
	\end{theorem}
	
	In summary, as expected, more precise approximation spaces are capable of reaching higher accuracy when sufficiently precise approximators are considered. However, increasing the precision of the approximation space also increases the size of the approximation space and thus the computational complexity. Therefore, the choice of approximation space and approximator should strike a balance between expressiveness and complexity. Note, however, that if $\AbS[1] \leqp \AbS[2]$, $\AbS[1]$ is essentially a subspace of $\AbS[2]$. Now, let $\aop[1] \in \appr[{\AbS[1]}]$ and  $\aop[2] \in \appr[{\AbS[2]}]$ such that $(\aop[1] \circ \gal) \leqp \aop[2]$. Then $\KK(\aop[1]) \in \AbS[2]$ and $\KK(\aop[1]) \leqp \KK(\aop[2])$. This entails that if we are not satisfied with the accuracy of $\aop[1]$, we do not need to start from scratch when calculating the Kripke-Kleene fixpoint with $\aop[2]$ in the more precise space. Instead, we can start from $\KK(\aop[1])$. Similarly, it is easy to verify that for every $\mathfrak{X}_1 \in \AbS[1]$, if $\mathfrak{X}_1$ is $\aop[1]$-reliable and -prudent, then $\mathfrak{X}_1 \in \AbS[2]$ is $\aop[2]$-reliable and -prudent. Hence, also for the well-founded model a complete restart is unnecessary. Consequently, instead of using a fixed approximation space, we can refine the space ad hoc. Essentially, this iterative procedure implicitly constructs a well-balanced approximation space.
	
	\section*{Conclusion}
	This paper proposes a conservative extension of consistent (interval-)\AFT by relaxing the constraints on the approximation space. With this extension the approximation space need not be intervals, more general approximants are allowed. The work is motivated by examples from different non-monotonic reasoning formalisms where standard \AFT fails. On the one hand, we see that standard \AFT can only handle problems in which the non-monotonic operator operates on a complete lattice. However, in many practical applications this is not the case. In this paper we gave a concrete example from the domain of abstract argumentation frameworks, but similar examples also exist in other domains. For example, a non-monotonic recursive function-definition in functional programming will often naturally obtain an operator on a bounded-complete cpo, but not necessarily on a complete lattice. On the other hand, \AFT sometimes lacks expressive power, as was indicated by an example from auto-epistemic logic. By allowing more general approximants, \AFT is capable of resolving the issues for both examples, showcasing that the extension improves both expressiveness and flexibility. We leave it for future work to investigate any formal results of just how far the improvements go. Additionally, by zooming in on the requirements of \AFT, the paper gives new insights on the key properties of \AFT. Finally, by introducing a new level of customization in the approximation space, the extension poses a new challenge. Since there are multiple possible approximation spaces, it is not necessarily immediately clear which approximation space is best suited for an application. A good approximation space for an application strikes a balance between expressive power and computational complexity. After examining the relations between approximation spaces, we see that we do not need to decide on a fixed approximation space from the beginning, instead we can iteratively refine our approximation space ad hoc, essentially resulting in a well-balanced approximation space.
	
	\bibliographystyle{kr}
	\bibliography{krrlib.bib}

\end{document}